\documentclass[letterpaper, 10 pt, conference]{ieeeconf}  

\IEEEoverridecommandlockouts                              

\usepackage{amsmath}
\usepackage{amssymb}
\usepackage{float}
\usepackage{graphicx}
\usepackage{wrapfig}
\usepackage{booktabs}
\usepackage{tabularx}
\usepackage{multirow}
\usepackage{array}
\usepackage{xcolor}
\let\labelindent\relax
\usepackage{enumitem}
\usepackage{xspace}
\usepackage{makecell}
\usepackage{threeparttable}
\usepackage{adjustbox}
\usepackage[table]{xcolor}
\usepackage{caption}
\usepackage{subcaption}
\usepackage{cite}
\usepackage{url}
\usepackage[breaklinks,colorlinks]{hyperref}
\usepackage{ragged2e}
\usepackage{cuted}
\usepackage{capt-of}
\usepackage{wrapfig}

\newcommand{\name}{ObjectWorld}

\title{\LARGE \bf \name: Scaling Object-Centric World Models with A Large-Scale 3D Full-Object Flow Dataset}

\author{Haonan Chen$^{1,3}$, Sicheng Isabella Lu$^{1,2}$, Haitian Li$^{1,2}$, Jiajin Wu$^{2}$, Hexiang Wei$^{2}$, Jiawei Zhang$^{1,2}$,\\ Xin Fei$^{1}$, Yiwen Hou$^{1}$, Chenrui Tie$^{1}$, Chongkai Gao$^{1}$, Sheng Wang$^{2}$, Lin Shao$^{\dagger1,2,3}$%
\thanks{$\dagger$ denotes the corresponding author. }
\thanks{$^{1}$School of Computing, National University of Singapore; $^{2}$RoboScience; $^{3}$NUS Guangzhou Research Translation and Innovation Institute}
}

\begin{document}

\maketitle
\thispagestyle{empty}
\pagestyle{empty}


\begin{strip}
\vspace{-18mm}
  \begin{center}
  \includegraphics[width=1.0\textwidth]{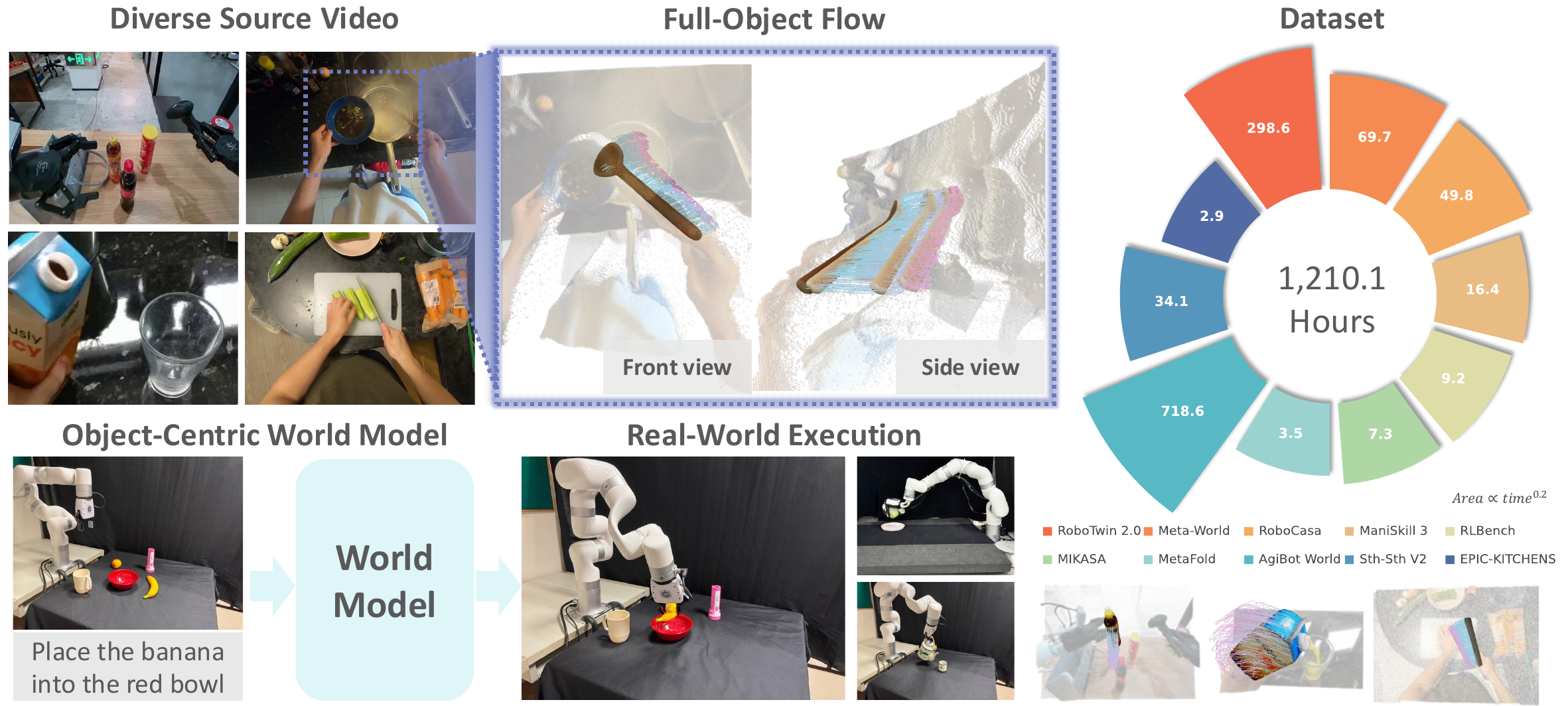}
  \end{center}
  \vspace{-3mm}
  \captionof{figure}{\textbf{From diverse videos to executable full-object flow.} We construct a 1,210.1-hour dataset of 3D full-object flow from diverse real-world and simulated videos. An object-centric world model predicts dense motion over the complete object surface from image-and-language inputs, and the predicted flow guides downstream robot execution. The radial chart summarizes its composition across seven simulation datasets and three large-scale real-world video sources.}
  \vspace{-4mm}
  \label{fig:teaser}
\end{strip}

\begin{abstract}

Generative world modeling from large-scale visual data offers a promising path toward scalable robot manipulation. This calls for an intermediate representation that can be supervised at scale and remains actionable for downstream control. We propose 3D full-object flow, an object-centric motion representation that assigns dense 3D trajectories to points across the complete surface of a target object, including its occluded regions. By extending visible-surface object flow to the full object geometry, this representation makes complete object motion explicit and applies uniformly to both rigid and deformable objects.
To realize the potential of this representation at scale, we introduce a 1210-hour full-object flow dataset that addresses the scarcity of large-scale supervision for complete 3D object motion. The dataset integrates diverse real-world videos and physical simulations, with a scalable annotation pipeline for recovering full-object motion from real-world observations and quality metrics for systematically assessing annotation quality.
We then train a world model that takes an image and a task description as input and predicts dense full-object flow over the reconstructed object geometry.  Experiments show reliable annotation quality, data-scalable flow prediction, and effective downstream manipulation across novel scenes and robot embodiments. Project website with additional results and videos: \href{https://nus-lins-lab.github.io/ObjectWorld/}{https://nus-lins-lab.github.io/ObjectWorld/}.

\end{abstract}


\section{Introduction}
\label{sec:intro}

Recent progress in robot foundation models and vision-language-action (VLA) models suggests that robotic manipulation can benefit substantially from scale~\cite{openvla, rt2, o2024open}. However, action-labeled robot data remain costly to collect, embodiment-dependent, and far less abundant than visual data~\cite{o2024open, grauman2022ego4d}. Generative world models therefore offer a promising path beyond direct action supervision by learning how objects and scenes evolve from visual experience~\cite{finn2017deep, pointworld, gao2025flip, wu2024ivideogpt, hou2026world}. Connecting these predictions to robot execution requires an intermediate representation that is both scalable to learn and actionable for downstream control~\cite{ebert2018visual, hou2026world}.

Prior representations each capture useful structure but expose trade-offs. Pixel-space predictions, such as goal images~\cite{susie} or videos~\cite{wu2024ivideogpt}, preserve rich visual information, but they are indirect for control and typically require an additional grounding step. Sparse keypoint or point-based interfaces represent manipulation objectives using landmarks or affordance points~\cite{huang2024rekep, moka}, but may miss the complete object motion. 6-DoF object poses offer metrically grounded state descriptions, but they rely on rigid-body assumptions and cannot capture non-rigid deformations~\cite{objectforesight, goalvla}.
Recent flow-based interfaces use object flow as an intermediate representation for manipulation~\cite{Im2flow2act, dream2flow, li2025novaflow}. However, they mainly model motion over visible or tracked object regions, leaving unobserved surfaces outside the motion field. We refer to object flow over visible regions as partial object flow.

To make object motion complete and actionable, we introduce 3D full-object flow, which specifies dense trajectories over the complete target-object surface as an explicit motion target for robot execution. Utilizing the reconstructed geometry of a target object, this representation maps dense trajectories over the entire object surface across time.
It makes complete object motion explicit in a unified 3D geometry, while remaining grounded in object kinematics rather than robot-specific actions. The resulting dense correspondence can describe rigid motion and non-rigid deformation, and provides an object-level motion target that downstream controllers can optimize against.

A key obstacle to adopting and scaling this complete representation is the scarcity of high-quality training data~\cite{firoozi2025foundation, gao2026vla, wang2026vision}. While numerous large-scale human video datasets and robotic trajectory collections exist~\cite{o2024open, grauman2022ego4d, damen2020epic, khazatsky2024droid}, their supervision signals are predominantly limited to semantic action labels, sparse 2D tracking, or robot-specific proprioception. They lack the dense, unified 3D full-object flow annotations required to supervise a generalizable object-centric world model across diverse tasks and scenes.

In this work, we address both the representation choice and the supervision bottleneck. We construct a 1210.1-hour 3D full-object flow dataset from diverse real-world videos and physical simulation. To build the real-world portion at scale, we introduce an automatic annotation pipeline that converts in-the-wild manipulation videos into 3D full-object flow, and we systematically assess the quality of the resulting annotations. On top of this dataset, we develop an object-centric world model framework that takes an image and a task description, reconstructs the initial geometry of the target object, 
and generates dense trajectories over the full object surface with a flow generator. The generated full-object flow is then consumed by a lightweight optimization-based execution module that translates object motion into end-effector trajectories.
Experiments validate the full pipeline from annotation to downstream execution, demonstrating reliable flow supervision, 
improved prediction through data scaling and model design,
and effective robot manipulation with transfer across environments and embodiments.

\noindent\textbf{Contributions.} 
\textbf{(i)} We formulate 3D full-object flow as an object-centric interface between generative world models and downstream robot execution.
\textbf{(ii)} We provide a 1210.1-hour 3D full-object flow dataset built from diverse real-world videos and physical simulation, together with a scalable annotation pipeline and quality assessment for the generated annotations.
\textbf{(iii)} We develop an object-centric world model framework that predicts dense full-object trajectories from image-and-language inputs.
\textbf{(iv)} We conduct extensive experiments across annotation, data-scale, prediction, and execution, showing effective robot manipulation and generalization to novel scenes and embodiments. 

\section{Related Works}
\label{sec:related}
\subsection{World Models for Robotic Manipulation}

World models have been increasingly used in robotic manipulation to predict future states or intermediate representations for planning and control~\cite{finn2017deep, ebert2018visual, hou2026world}. A common strategy is to use intermediate representations less tied to a specific action space or embodiment, such as future images~\cite{susie}, videos~\cite{wu2024ivideogpt}, or keypoints~\cite{huang2024rekep, moka}. These abstractions can improve transfer across tasks and embodiments, but are often indirect for control or sparse in object coverage.

Recent work has begun to model object motion more directly. Pose-based methods use 6-DoF object poses to represent future object states or task goals~\cite{objectforesight, goalvla, videomanip}, providing metric structure for rigid-object manipulation but not naturally capturing non-rigid deformation. Im2Flow2Act~\cite{Im2flow2act} uses 2D object flow over tracked image points, Dream2Flow~\cite{dream2flow} and NovaFlow~\cite{li2025novaflow} derive object-centric 3D flow from masked tracked points in generated videos, and PointWorld~\cite{pointworld} predicts action-conditioned full-scene 3D point flow from RGB-D observations and robot actions. 
These works demonstrate the value of flow-based interfaces, while their prediction targets are typically defined on image tracks, masked visible regions, or partially observed scene geometry. Our work predicts full-surface 3D trajectories over a reconstructed complete target object for downstream optimization.

\subsection{Datasets and Scalable Supervision for 3D Object Flow}

Scalable supervised learning of dense object-centric 3D motion prediction benefits from large-scale 3D motion annotations, yet such supervision remains limited in real-world data~\cite{pointworld, firoozi2025foundation, wang2026vision}.  Large video datasets~\cite{damen2020epic, goyal2017something, mahdisoltani2018effectiveness, grauman2022ego4d} provide coverage of human-object interaction, but their annotations do not provide dense 3D trajectories of manipulated objects. Datasets with richer 3D annotations, such as HOI4D~\cite{liu2022hoi4d}, DexYCB~\cite{chao2021dexycb}, ARCTIC~\cite{fan2023arctic}, and HOT3D~\cite{hot3d}, provide object poses, meshes, contacts, or tracking signals. However, their reliance on curated capture setups, markers, or labor-intensive annotation limits their scalability for dense object-flow supervision. Large robot datasets, such as Open X-Embodiment~\cite{o2024open}, DROID~\cite{khazatsky2024droid}, and AgiBot World~\cite{bu2025agibot}, further scale real-world manipulation data, but they primarily provide robot observation-action trajectories rather than unified 3D full object flow annotations. To address this gap, we build a scalable annotation pipeline that converts heterogeneous real-world videos into 3D full object flow supervision, emphasizing dense target object motion rather than action labels, sparse poses, or robot-specific trajectories.

\section{Method}
\label{sec:method}

We first formulate object-centric world modeling and flow-conditioned robot execution in Sec.~\ref{sec:formulation}. We then describe the annotation pipeline and dataset construction in Sec.~\ref{sec:data_pipeline}, the world model architecture and training strategy in Sec.~\ref{sec:model}, and the downstream action optimization in Sec.~\ref{sec:action_optimization}.

\begin{figure*}[t]
  \centering
  \includegraphics[width=1.0\textwidth]{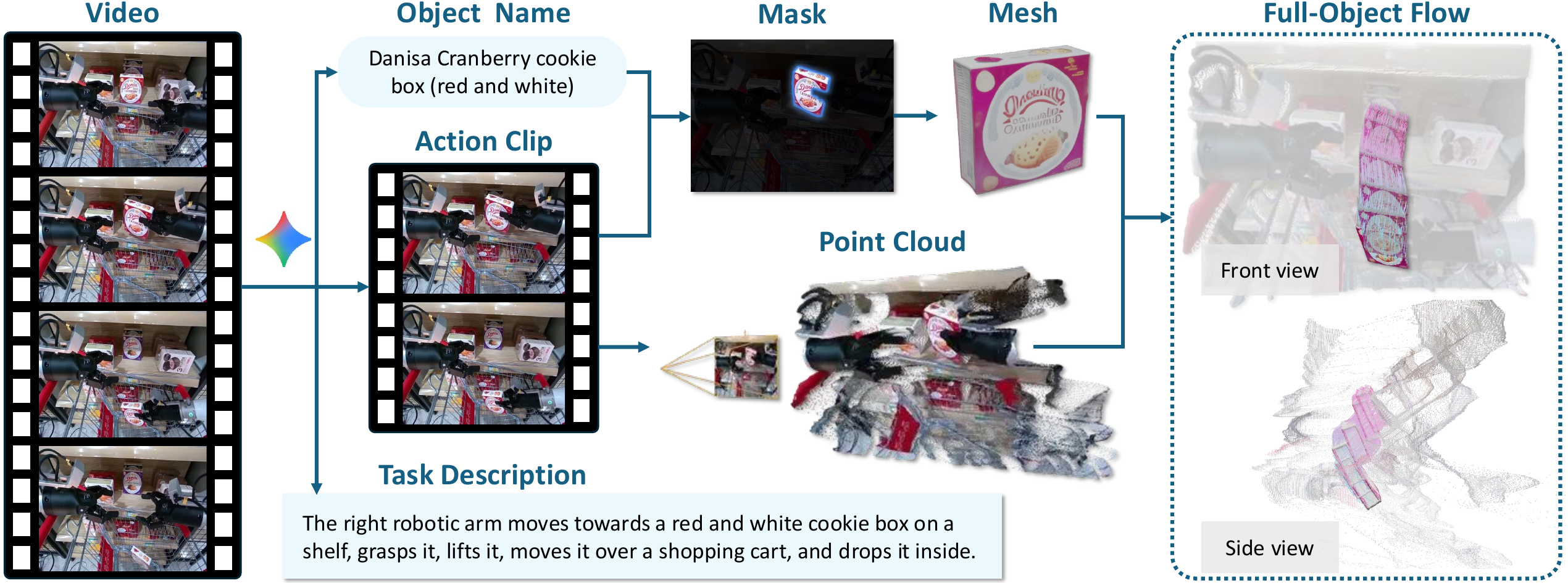}
  \vspace{-15pt}
  \caption{\textbf{Automatic data annotation pipeline.} Given an in-the-wild manipulation video, a VLM identifies the target object, describes the task, and segments the video into atomic action clips. For each clip, our pipeline extracts the target-object masks, reconstructs the complete object mesh and the point clouds, and aligns the mesh with the point clouds across frames to recover the object motion. The  motion is propagated over the complete object surface to produce 3D full-object flow.}
  \vspace{-15pt}
  \label{fig:data_pipeline}
\end{figure*}

\subsection{Problem Formulation}
\label{sec:formulation}

Given an initial RGBD observation $\mathcal{I}$ and a natural language instruction $\mathcal{L}$, our world model predicts a 3D full-object flow
$\hat{\mathcal{P}}^{\mathrm{obj}}_{1:T}=\{\hat{\mathbf{p}}_{i,t}\in\mathbb{R}^{3}\}_{i=1,t=1}^{N^\mathrm{obj},T}$,
where $N^\mathrm{obj}=512$ is the number of sampled object points and $T=48$ is the prediction horizon. Each $\hat{\mathbf{p}}_{i,t}$ denotes the predicted 3D position of object point $i$ at time step $t$. This representation specifies the desired object motion for accomplishing the task, and serves as an interface between high-level task prediction and low-level robot execution.
Conditioned on the predicted full-object flow, the robot executes the task by solving a flow-conditioned action optimization problem:
\[
\boldsymbol{\tau}^{\star}
=
\arg\min_{\boldsymbol{\tau}\in\mathcal{A}}
\mathcal{D}
\left(
\Phi(\mathcal{I}, \boldsymbol{\tau}),
\hat{\mathcal{P}}^{\mathrm{obj}}_{1:T}
\right)
+
\lambda \mathcal{R}(\boldsymbol{\tau}),
\]
where $\boldsymbol{\tau}$ denotes the robot action variable, instantiated as an end-effector trajectory in our implementation; $\mathcal{A}$ is the feasible trajectory space; $\Phi(\mathcal{I},\boldsymbol{\tau})$ denotes the object motion induced by executing $\boldsymbol{\tau}$ from the initial observation; $\mathcal{D}$ measures its discrepancy to the predicted full-object flow; and $\mathcal{R}$ regularizes action feasibility and smoothness.

\subsection{Dataset Curation and Pipeline}
\label{sec:data_pipeline}

\subsubsection{Annotation of Real-World Videos}

To enable scalable supervision, we build an annotation pipeline that converts heterogeneous in-the-wild videos into 3D full-object flow, as illustrated in Fig.~\ref{fig:data_pipeline}. The pipeline does not require robot actions, depth measurements, calibrated cameras, or pre-scanned object meshes. It first uses Gemini 2.5 Pro~\cite{gemini2.5} as a Vision-Language Model (VLM) to identify the target object, generate a task description, and segment each raw video into atomic manipulation clips, each corresponding to a single contact-and-separation interaction between the actor and the target object. For each clip, we use Track4World~\cite{track4world} with its Depth Anything 3 (DA3) backbone~\cite{depthanything3} to estimate camera poses and reconstruct per-frame point clouds in a common coordinate system anchored to the first-frame camera.

We then localize the target object using VLM-based object proposals followed by SeC~\cite{sec} for mask propagation. Given the propagated target masks, SAM 3D~\cite{sam3d} reconstructs the complete object mesh, Any6D~\cite{any6d} estimates its initial pose, and FoundationPose~\cite{wen2024foundationpose} tracks its pose throughout the clip. To obtain full-object flow, we uniformly sample points from the reconstructed mesh in its canonical object frame and, at each timestep, transform them by composing the tracked object pose with the corresponding camera-to-reference transformation recovered by Track4World. The resulting dense trajectories, expressed in the first-frame camera coordinate system, constitute the full-object flow $\mathcal{P}^{\mathrm{obj}}_{1:T}$.

\subsubsection{Data Filtering}

To filter low-quality annotations, we evaluate the spatial and temporal consistency of the estimated object motion. At each frame, we project the reconstructed object under its estimated pose into the image plane and compute the IoU between the projected object region and the propagated target mask. We aggregate the IoU scores over the clip to identify annotations with unstable pose tracking.

We manually curate a validation set of expert-labeled examples to calibrate the filtering criterion. We evaluate the consistency metric on this set and select a threshold that achieves high precision while retaining sufficient coverage according to the expert labels. We then apply the threshold across the dataset to remove low-quality annotations.

\begin{figure*}[t]
  \centering
  \includegraphics[width=1.0\textwidth]{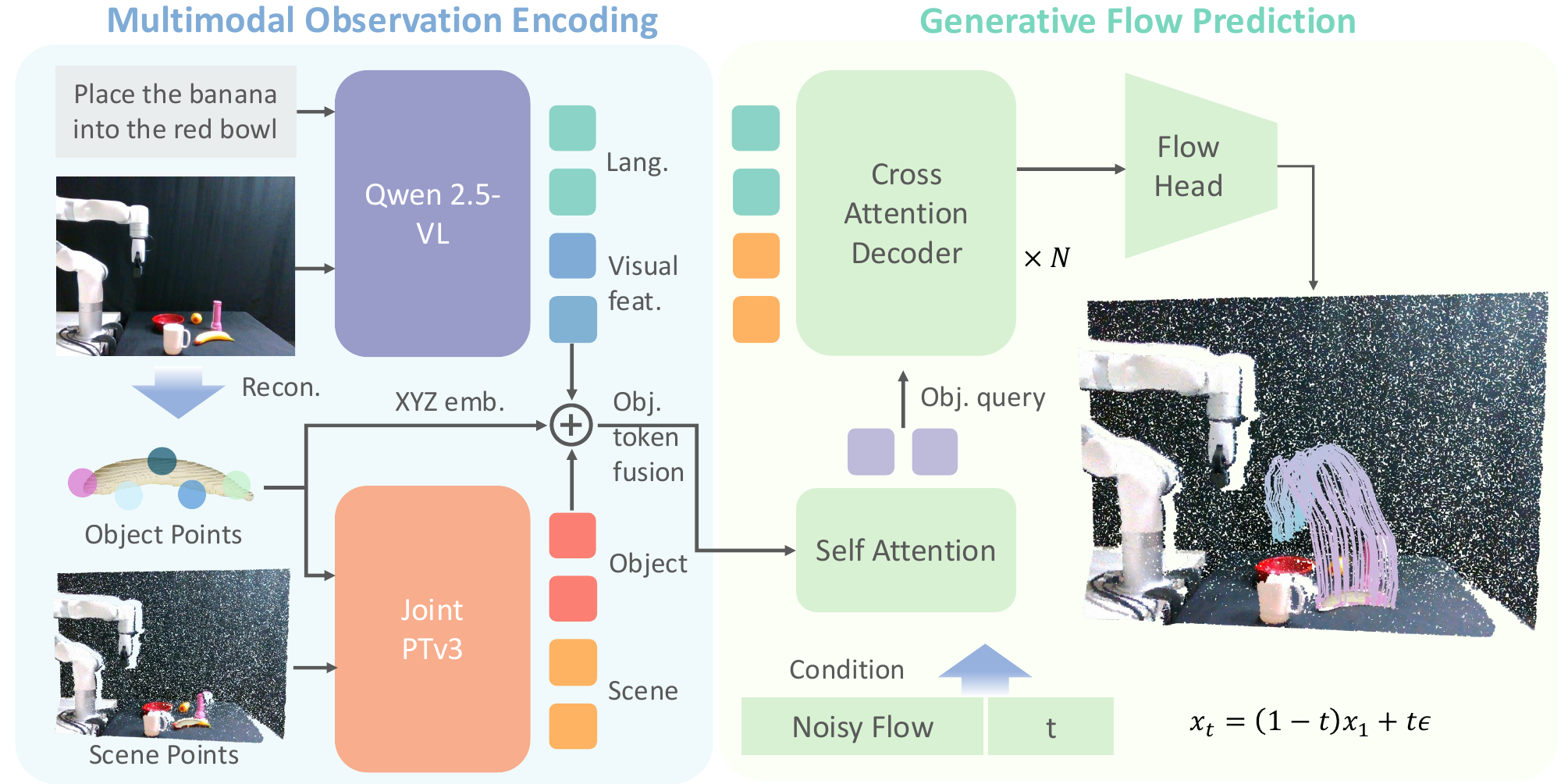}
  \vspace{-15pt}
  \caption{\textbf{Object-centric full-object flow prediction framework.} Given an initial RGB-D observation and task instruction, Qwen2.5-VL~\cite{bai2025qwen25vltechnicalreport} extracts task-conditioned visual-language features, while a joint PointTransformer V3~\cite{ptv3} (PTv3) encodes reconstructed object and scene geometry. These features are fused into object tokens
  and refined by self-attention. The object queries attend to scene and language context through a cross-attention decoder, conditioned on noisy flow and time $t$, before a flow head predicts dense trajectories over the complete object surface.}
  \vspace{-15pt}
  \label{fig:model}
\end{figure*}

\subsubsection{Dataset Composition}

Our dataset comprises 1,210.1 hours of manipulation data measured at a standardized 16 FPS: 454.5 hours of simulated trajectories and 755.6 hours of automatically annotated real-world videos.

For the simulation subset, we aggregate 198,947 trajectories spanning 251 tasks from seven datasets: ManiSkill 3~\cite{tao2024maniskill3}, MetaFold~\cite{chen2025metafold}, Meta-World~\cite{yu2020metaworld}, MIKASA~\cite{mikasa}, RLBench~\cite{james2020rlbench}, RoboCasa~\cite{nasiriany2024robocasa}, and RoboTwin 2.0~\cite{chen2025robotwin2}. MetaFold contributes deformable-object manipulation data.

For real-world data, we collect videos with diverse viewpoints and embodiments from three sources: robot demonstrations from AgiBot World~\cite{bu2025agibot}, third-person human demonstrations from Something-Something V2~\cite{goyal2017something}, and egocentric human demonstrations from EPIC-KITCHENS~\cite{damen2020epic}. After automatic annotation and quality filtering, we retain 289,976 clips totaling 755.6 hours.



\subsection{Full-Object Flow Prediction}
\label{sec:model}

Given an initial observation and a task instruction, we predict the future motion of the complete target-object surface, as illustrated in Fig.~\ref{fig:model}. Let $\mathbf{P}^{o}\in\mathbb{R}^{N_o\times 3}$ denote the initial object surface and
$\mathbf{P}^{s}\in\mathbb{R}^{N_s\times 3}$ the scene geometry.
Our predictor outputs a dense displacement trajectory
\begin{equation}
    \mathbf{F}
    =
    \{\Delta \mathbf{P}^{o}_{t}\}_{t=1}^{T}
    \in \mathbb{R}^{T\times N_o\times 3},
\end{equation}
where each displacement is defined with respect to the initial object surface.
This object-centric representation describes how the manipulated object evolves independently of the robot embodiment.

\paragraph{Initial 3D observation construction}
We construct an object-centric 3D observation from the first RGB-D frame.
We first obtain the target-object mask using SAM2~\cite{sam2} and reconstruct its complete geometry with SAM3D~\cite{sam3d}.
Any6D~\cite{any6d} estimates the initial object pose, which places the reconstructed mesh into the observed scene.
We then sample points on the transformed mesh surface to obtain the complete object point set $\mathbf{P}^{o}$.
In parallel, the scene point cloud $\mathbf{P}^{s}$ is obtained by back-projecting the observed depth map using the camera intrinsics.
Thus, the model receives both a complete object-centric geometric representation and the
observed scene geometry from the initial frame.

\paragraph{Multimodal observation encoding}
We encode semantic and geometric information through complementary visual-language and 3D pathways.
A vision-language encoder processes the initial RGB image together with the task instruction and produces task-conditioned visual features and language tokens.
For each object point, we project its 3D coordinate into the image and sample the corresponding visual feature, providing a point-wise semantic representation.

In parallel, we jointly encode the object and scene point clouds using Point Transformer V3 (PTv3)~\cite{ptv3}.
Joint 3D encoding allows the object representation to incorporate its surrounding geometric context.
For each object point $\mathbf{p}^{o}_{i}$, we construct an object token by combining its spatial embedding, point-wise visual-language feature, and 3D feature:
\begin{equation}
    \mathbf{h}^{o}_{i}
    =
    \phi_{\mathrm{xyz}}(\mathbf{p}^{o}_{i})
    +
    \phi_{\mathrm{vlm}}(\mathbf{f}^{\mathrm{vlm}}_{i})
    +
    \phi_{\mathrm{3D}}(\mathbf{f}^{\mathrm{3D}}_{i}).
\end{equation}
The scene-side PTv3 features are retained as contextual scene tokens, while the language tokens preserve the task-level semantic information.


\paragraph{Generative full-object flow prediction}
We formulate future-flow prediction using an $x_1$-prediction parameterization
over a linear noising path.
Let $\mathbf{x}_1$ denote the full-object trajectory and
$\mathbf{x}_0\sim\mathcal{N}(\mathbf{0},\mathbf{I})$ denote Gaussian noise.
During training, we sample $t\sim\mathcal{U}(0,1)$ and construct
\begin{equation}
    \mathbf{x}_t
    =
    (1-t)\mathbf{x}_1
    +
    t\mathbf{x}_0.
\end{equation}

The noisy trajectory $\mathbf{x}_t$ is projected into the object tokens, while
the timestep $t$ modulates them through adaptive normalization.
The conditioned object tokens are refined by self-attention and then used as
queries in a cross-attention decoder, whose keys and values consist of the
scene and language tokens.
This allows each object point to reason jointly about the task semantics,
surrounding geometry, and the motion of the complete object surface.

The network directly predicts the clean trajectory,
\begin{equation}
    \hat{\mathbf{x}}_1
    =
    f_{\theta}
    \left(
        \mathbf{x}_t,
        t,
        \mathbf{H}^{o},
        \mathbf{H}^{s},
        \mathbf{H}^{l}
    \right),
\end{equation}
where $\mathbf{H}^{o}$, $\mathbf{H}^{s}$, and $\mathbf{H}^{l}$ denote the
object, scene, and language representations, respectively.
All $T$ future frames are predicted jointly.
We optimize
\begin{equation}
    \mathcal{L}_{\mathrm{flow}}
    =
    \left\|
        \hat{\mathbf{x}}_1-\mathbf{x}_1
    \right\|_2^2.
\end{equation}
At inference time, we initialize the trajectory from Gaussian noise and apply
the same $x_1$-prediction update for one or more refinement steps to obtain the full-object trajectory.

\paragraph{Training with Heterogeneous Data}
We train the flow predictor on heterogeneous simulated and real-world data, which introduces inconsistencies in coordinate orientation and temporal scale. To standardize orientation, we estimate gravity with GeoCalib~\cite{veicht2024geocalib} and rotate each scene such that the $z$-axis points opposite to gravity, applying the same transformation to both 3D points and object-flow trajectories. This normalization provides a consistent gravity-aligned coordinate frame across datasets, which is important because manipulation dynamics depend on gravity~\cite{videomanip}. To normalize temporal scale, we uniformly resample each action segment into $T$ frames, such that object flow represents spatial evolution over normalized action progress rather than absolute execution time.

\subsection{Flow-Conditioned Action Optimization}
\label{sec:action_optimization}


We execute the predicted full-object flow with a lightweight local patch fitting optimizer. A preceding grasp module provides the initial manipulator pose
$\mathbf{X}^{\mathrm{m}}_1=(\mathbf{R}^{\mathrm{m}}_1,
\mathbf{t}^{\mathrm{m}}_1)$ and the contact region. We use sample-based grasping
with a parallel gripper and T(R,O) Grasp~\cite{tro} for the dexterous hand. Let
$\mathcal{C}\subseteq\{1,\ldots,N^{\mathrm{obj}}\}$ denote a local contact
patch. We express its points in the initial manipulator frame as
\begin{equation}
\mathbf{y}_i =
(\mathbf{R}^{\mathrm{m}}_1)^\top
(\mathbf{p}_{i,1}-\mathbf{t}^{\mathrm{m}}_1),
\quad i\in\mathcal{C}.
\end{equation}
The local coordinates $\mathbf{y}_i$ are kept fixed when recovering the trajectory. This imposes only a local rigidity assumption around the contact patch, rather than a global rigidity assumption on the object; the predicted full-object flow can still represent non-rigid motion over the rest of the surface.

At each time step, we recover the manipulator pose by robustly aligning this
local patch with its predicted positions:
\begin{equation}
(\mathbf{R}^{\mathrm{m}}_t,\mathbf{t}^{\mathrm{m}}_t)
=
\arg\min_{\mathbf{R}\in SO(3),\,\mathbf{t}\in\mathbb{R}^{3}}
\sum_{i\in\mathcal{C}}
\rho_{\delta}
\left(
\left\|\mathbf{R}\mathbf{y}_i+\mathbf{t}
-\hat{\mathbf{p}}_{i,t}\right\|_2
\right).
\end{equation}
The resulting poses form the execution trajectory
$\boldsymbol{\tau}=\{\mathbf{X}^{\mathrm{m}}_t\}_{t=1}^{T}$.
We solve the Huber-robust objective using IRLS, with each subproblem given by
weighted Kabsch alignment.

\section{Experiments}
\label{sec:exp}

Our experiments address the following key questions:

\begin{enumerate}[label=\textbf{Q\arabic*:}, leftmargin=*, itemsep=0pt, topsep=2pt, parsep=0pt, partopsep=0pt]
    \item How reliable are the full-object flow annotations produced and retained by our large-scale pipeline?
    
    \item How accurately does our model predict full-object flow?
    
    \item How do data scale and model design affect prediction?
    
    \item Can full-object flow serve as an effective
    motion target for downstream robot manipulation?
    
    \item How well does our system generalize across different environments and robot embodiments?
\end{enumerate}


\begin{table}[h]
\vspace{-4pt}
\centering

\caption{Experiment map. A: annotation validation; B: flow prediction and scaling; C: robot execution and generalization.}
\vspace{-6pt}
\label{tab:exp_map}

\begingroup
\footnotesize
\setlength{\tabcolsep}{3pt}
\renewcommand{\arraystretch}{0.94}

\begin{tabularx}{\linewidth}{
    @{}
    >{\RaggedRight\arraybackslash}p{0.07\linewidth}
    >{\RaggedRight\arraybackslash}X
    >{\centering\arraybackslash}p{0.13\linewidth}
    @{}
}
\toprule
\textbf{ID} & \textbf{Study} & \textbf{Supports} \\
\midrule

A1 & Pipeline validation against simulator ground truth
   & Q1 \\

A2 & Filter-gate validation against expert labels
   & Q1 \\

A3 & Full-object vs.\ partial-object flow annotation
   & Q1 \\

\midrule

B1 & Flow predictor comparison with baselines
   & Q2 \\


B2 & Robustness of flow predictor & Q2 \\

B3 & Data scaling experiment
   & Q3 \\
   
B4 & Task-language ablation study & Q3 \\

\midrule

C1 & Simulation execution with predicted flow
   & Q4, Q5 \\

C2 & Real-world execution with predicted flow
   & Q4, Q5 \\



\bottomrule
\end{tabularx}
\endgroup

\vspace{-10pt}
\end{table}
\vspace{-5pt}


\begin{table*}[t]
\centering
\vspace{2pt}
\caption{\textbf{A1. Ground-truth validation and runtime of the annotation
pipeline on simulation.}
Quality metrics are medians over 97 successfully annotated RoboTwin clips.
$D$ denotes the ground-truth object diameter.
Times are measured on standardized 3-second clips.
$^\dagger$Amortized time per call; $^\ddagger$CPU flow construction.}
\label{tab:annotation_validation}

\begingroup
\small
\setlength{\tabcolsep}{2.5pt}
\renewcommand{\arraystretch}{1.05}

\begin{tabularx}{\textwidth}{
    @{}
    >{\RaggedRight\arraybackslash}p{0.205\textwidth}
    >{\RaggedRight\arraybackslash}p{0.155\textwidth}
    >{\RaggedRight\arraybackslash}p{0.170\textwidth}
    >{\RaggedRight\arraybackslash}X
    >{\centering\arraybackslash}p{0.120\textwidth}
    >{\centering\arraybackslash}p{0.090\textwidth}
    @{}
}
\toprule
\textbf{Module / Stage}
& \textbf{Evaluated Output}
& \textbf{Protocol}
& \textbf{Metric}
& \textbf{Median}
& \textbf{Time (s)} \\
\midrule

Gemini 2.5 Pro (VLM)
& Localization
& Against GT box
& Box IoU $\uparrow$
& \textbf{0.908}
& 12.04$^\dagger$ \\

SeC
& Mask propagation
& Against GT masks
& Mask IoU $\uparrow$
& \textbf{0.873}
& 11.20 \\

Track4World (DA3)
& Relative depth
& Median-scale aligned
& AbsRel $\downarrow$ / $\delta<1.25$ $\uparrow$
& \textbf{0.089 / 0.826}
& 8.25 \\

\addlinespace[1pt]
\multirow{3}{*}{SAM 3D Objects}
& Complete geometry
& Shape aligned
& CD/$D$ $\downarrow$
& \textbf{2.61\%}
& \multirow{3}{*}{7.41} \\

& Hidden geometry
& Visible-only Sim(3)
& Vis./hid.\ surface error/$D$ $\downarrow$
& \textbf{4.22/7.28\%}
& \\

& Hidden coverage
& Visible-only Sim(3)
& Recall@$5\%D$ $\uparrow$
& \textbf{0.462}
& \\
\addlinespace[1pt]

Any6D + FoundationPose
& Surface motion
& Initial-frame Sim(3)
& Surface-trajectory error/$D$ $\downarrow$
& \textbf{11.50\%}
& 5.17+3.53 \\

3D full-object flow
& Dense trajectories
& Frozen correspondence
& EPE [cm] / EPE/$D$ $\downarrow$
& \textbf{3.71 / 26.11\%}
& 0.19$^\ddagger$ \\

Reprojection-IoU filter
& Quality selection
& Golden vs.\ rejected
& Frozen-flow EPE/$D$ $\downarrow$
& \textbf{16.29/35.49\%}
& -- \\

\bottomrule
\end{tabularx}
\endgroup
\vspace{-4mm}
\end{table*}

\subsection{Experimental Setup}
\label{sec:experimental_setup}

\noindent\textbf{Evaluation strategy.}
We organize the evaluation around the five questions above and group the
studies into three families: annotation validation (Q1), flow prediction and data scaling and ablation (Q2--Q3), and robot execution and generalization (Q4--Q5), as
summarized in Table~\ref{tab:exp_map}.



\noindent\textbf{Platforms and embodiments.}
We evaluate simulation performance in ManiSkill 3~\cite{tao2024maniskill3}
and LIBERO~\cite{liu2023libero}. Real-world evaluation covers two settings, with parallel-gripper and dexterous-hand systems performing different tasks in their respective environments.

\noindent\textbf{Training data.}
Using the filtering protocol validated in Fig.~\ref{fig:filter_confusion}, we
retain 288.7 hours of real-world videos as the Golden subset. Unless otherwise
specified, our method and the learning-based baselines are trained on all
simulation data together with this Golden subset.

\subsection{Annotation Pipeline Validation (Q1)}
\label{sec:exp_annotation}

We evaluate the reliability of our annotation pipeline through three complementary experiments: validation against simulator ground truth, validation of the filtering gates against expert judgments, and a comparison between full-object and partial-object flow annotations. 

\noindent\textbf{Validation against simulator ground truth.} We evaluate each annotation stage on 99 eligible 3-second RoboTwin clips. The pipeline successfully processes 97 clips, yielding a 98.0\% annotation rate. As shown in Table~\ref{tab:annotation_validation}, target localization and mask propagation achieve median IoUs of 0.908 and 0.873, respectively. Complete-object reconstruction obtains a Chamfer distance of 2.61\%$D$, with a hidden-surface error of 7.28\%$D$ under visible-only alignment. After pose estimation and tracking, the recovered surface motion has a trajectory error of 11.50\%$D$, and the final dense full-object flow achieves an EPE of 3.71\,cm (26.11\%$D$). These results indicate that the successive pipeline stages recover consistent object geometry and motion from monocular videos. The local GPU stages require approximately 35.6 seconds per clip and can be parallelized across clips. 

\begin{figure}[t]
    \centering
    \includegraphics[width=\columnwidth]{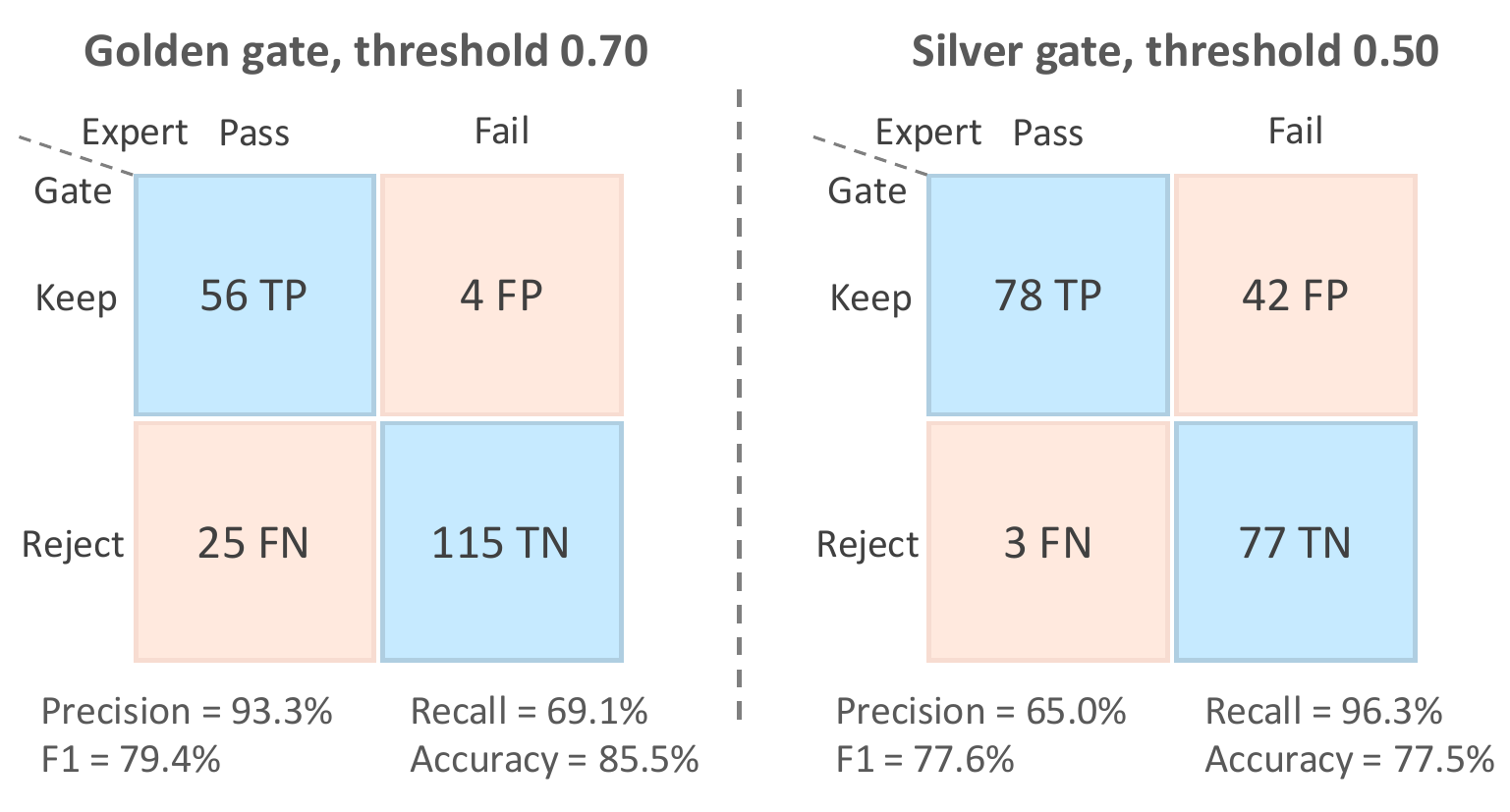}
    \caption{\textbf{A2. Expert validation of the filtering gates.}
Confusion matrices for the Golden and Silver gates on 200 expert-labeled
real-world clips.}
    \vspace{-10pt}
    \label{fig:filter_confusion}
\end{figure}

\begin{table}[t]
\centering
\caption{\textbf{A3. Full-object versus partial motion supervision.}
All results are medians. EPE is evaluated on identical GT material points;
Coverage@$5\%D$ is measured over the complete GT surface, with unsuccessful
cases counted as 0.}
\vspace{-5pt}
\label{tab:full_partial_annotation}

\begingroup
\small
\setlength{\tabcolsep}{3.5pt}
\renewcommand{\arraystretch}{1.05}

\begin{tabular}{@{}lccc@{}}
\toprule
\textbf{Representation}
& \textbf{EPE/$D$ $\downarrow$}
& \textbf{Surface Cov. $\uparrow$}
& \textbf{Hidden Cov. $\uparrow$} \\
\midrule
Track4World partial
& 37.22\%
& 28.4\%
& 9.8\% \\
Full-object flow
& \textbf{30.74\%}
& \textbf{45.8\%}
& \textbf{26.6\%} \\
\bottomrule
\end{tabular}

\endgroup
\vspace{-16pt}
\end{table}

\begin{table*}[t]
\centering
\vspace{2pt}
\caption{\textbf{Flow predictor evaluation.}
EPE is reported in centimeters, and Acc@$\,\tau$ denotes per-point accuracy
(\%). Subscript $T$ denotes the final frame.
$^\dagger$im2Flow2Act reports lateral 2D-flow error.}
\vspace{-4pt}
\label{tab:b1_b2_predictor_eval}

\begingroup
\footnotesize
\renewcommand{\arraystretch}{1.05}

\begin{subtable}[t]{0.600\linewidth}
\vspace{0pt}
\centering
\caption{\textbf{B1. Comparison with flow-prediction baselines.}}
\vspace{-3pt}
\label{tab:b1_flow_baselines}

\setlength{\tabcolsep}{2.5pt}
\begin{tabular*}{\linewidth}{
    @{\extracolsep{\fill}}lcccccc@{}
}
\toprule
\multirow{2}{*}{Method}
& \multicolumn{3}{c}{All Frames}
& \multicolumn{3}{c}{Final Frame} \\[-3pt]
\cmidrule(lr){2-4}\cmidrule(l){5-7}
& EPE$\downarrow$
& Acc@5$\uparrow$
& Acc@10$\uparrow$
& EPE$\downarrow$
& Acc@5$\uparrow$
& Acc@10$\uparrow$ \\[-2pt]
\midrule
ObjectForesight
& 10.24 & 48.9 & 64.3
& 21.18 & 9.9 & 25.9 \\
TraceGen
& 12.97 & 42.2 & 54.2
& 26.43 & 3.3 & 11.1 \\
im2Flow2Act$^\dagger$
& 12.58 & 36.7 & 63.2
& 20.88 & 18.8 & 38.0 \\
\textbf{Ours}
& \textbf{6.22} & \textbf{61.0} & \textbf{79.3}
& \textbf{11.17} & \textbf{28.6} & \textbf{60.1} \\
\bottomrule
\end{tabular*}
\end{subtable}%
\hspace{0.015\linewidth}%
\begin{subtable}[t]{0.38\linewidth}
\vspace{0pt}
\centering
\caption{\textbf{B2. Robustness to object-geometry noise.}}
\vspace{-3pt}
\label{tab:b2_geometry_noise}

\renewcommand{\arraystretch}{1.05}
\setlength{\tabcolsep}{2pt}
\begin{tabular*}{\linewidth}{
    @{\extracolsep{\fill}}ccccc@{}
}
\toprule
$\sigma$ (cm)
& EPE$_{\mathrm{all}}\downarrow$
& EPE$_T\downarrow$
& Acc@5$_T\uparrow$
& Acc@10$_T\uparrow$ \\
\midrule
0   & 6.22 & 11.17 & 28.6 & 60.1 \\
0.5 & 6.23 & 11.17 & 28.8 & 60.0 \\
1   & 6.23 & 11.17 & 28.9 & 60.2 \\
2   & 6.23 & 11.22 & 28.8 & 60.0 \\
5   & 6.34 & 11.61 & 26.6 & 58.3 \\
\bottomrule
\end{tabular*}
\end{subtable}

\endgroup
\vspace{-6pt}
\end{table*}

\noindent\textbf{Filtering against expert judgments.} We examine whether the filtering criterion can distinguish reliable annotations from failure cases. Under simulator ground truth, annotations retained by the Golden gate have a substantially lower median flow error than rejected annotations (16.29\%$D$ versus 35.49\%$D$). We further evaluate the gates on 200 expert-labeled real-world clips. As shown in Fig.~\ref{fig:filter_confusion}, the Golden gate achieves 93.3\% precision, while the Silver gate achieves 96.3\% recall. The two thresholds provide complementary operating points: the Golden gate prioritizes annotation quality, whereas the Silver gate retains broader coverage. 

\noindent\textbf{Full-object versus partial motion supervision.}
In Table~\ref{tab:full_partial_annotation}, we compare our full-object annotations with Track4World trajectories anchored
to the initially visible object surface. On identical GT material points,
full-object flow reduces median EPE from 37.22\%$D$ to 30.74\%$D$.
It also increases complete-surface and hidden-surface Coverage@$5\%D$ from
28.4\% to 45.8\% and from 9.8\% to 26.6\%, respectively. These results show
that full-object flow provides more accurate median motion while extending
valid supervision beyond the initially observed surface.

\subsection{Full-Object Flow Prediction (Q2)}
\label{sec:exp_prediction}

\noindent\textbf{Flow-prediction baseline comparison.}
Table~\ref{tab:b1_b2_predictor_eval}(\subref{tab:b1_flow_baselines})~compares
our model with the partial-flow prediction methods TraceGen~\cite{lee2026tracegen} and im2Flow2Act~\cite{Im2flow2act},
as well as ObjectForesight~\cite{objectforesight}, which models rigid-object motion. In contrast, our model predicts dense trajectories over the complete object surface and accommodates both rigid and
non-rigid motion. It achieves the best results across all all-frame and
final-frame metrics, with EPEs of 6.22\,cm and 11.17\,cm, respectively.
This advantage shows that our model more accurately captures
intermediate motion and long-horizon displacement.

\noindent\textbf{Robustness to object-geometry noise.} We further perturb the reconstructed object points with i.i.d.\ Gaussian noise while keeping the scene geometry and target flow unchanged. Table~\ref{tab:b1_b2_predictor_eval}(\subref{tab:b2_geometry_noise})~shows negligible degradation for noise levels up to 2\,cm. Even with 5\,cm noise, final-frame EPE remains 11.61\,cm and Acc@10 decreases only from 60.1\% to 58.3\%. Thus, the predictor is robust to moderate errors in the reconstructed object geometry.

\subsection{Data Scaling and Ablation Study (Q3)}
\label{sec:exp_scaling}

\begin{figure*}[t]
  \centering
  \includegraphics[width=1.0\textwidth]{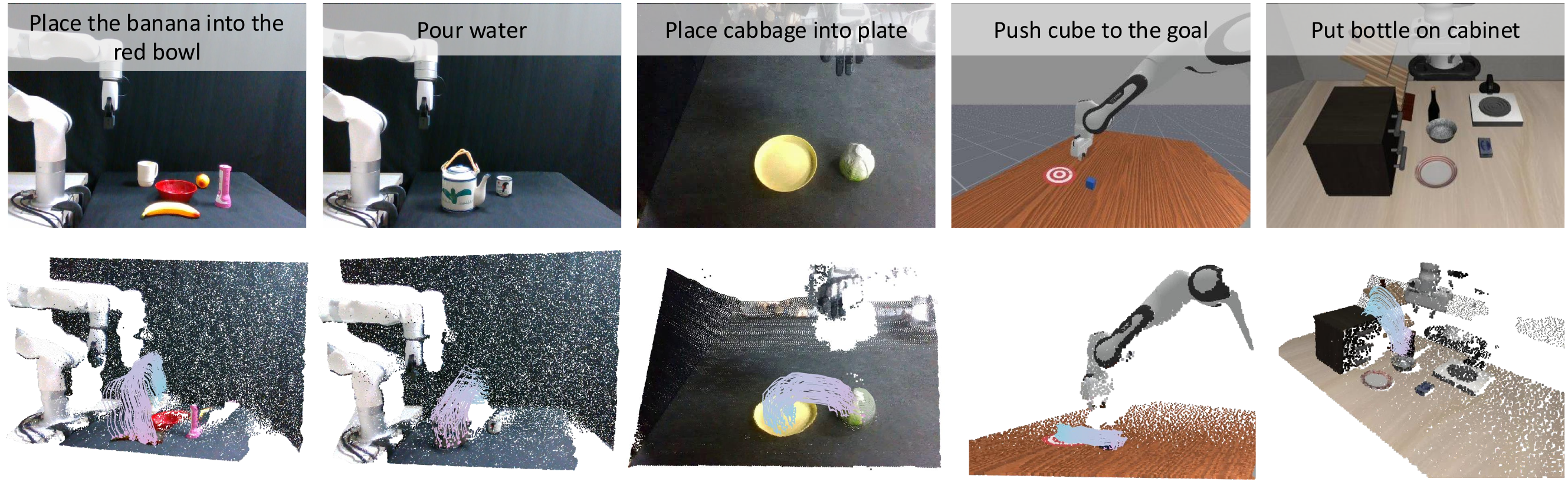}
  \vspace{-15pt}
\caption{\textbf{Qualitative full-object flow predictions.}
Given an initial RGB observation and task instruction (top), our model predicts dense 3D trajectories over the complete target-object surface (bottom) across real-world and simulated tasks.}

  \vspace{-15pt}
  \label{fig:sim_and_real}
\end{figure*}

\begin{table}[h]
\centering
\caption{
\textbf{B4. Task-language ablation.} Final-frame EPE.
}
\vspace{-5pt}
\label{tab:language_ablation}

\begingroup
\small
\renewcommand{\arraystretch}{0.9}

\begin{tabular*}{\columnwidth}{
    @{\extracolsep{\fill}}
    lccc
    @{}
}
\toprule
\textbf{Caption}
& \textbf{AgiBot}
& \textbf{SthV2}
& \textbf{Overall} \\
\midrule
Placeholder
& 16.39 & 17.48 & 13.76 \\
Task-specific
& \textbf{14.86} & \textbf{15.27} & \textbf{13.01} \\
\midrule
$\Delta$
& $-1.53$ & $-2.21$ & $-0.75$ \\
\bottomrule
\end{tabular*}
\vspace{-8pt}
\endgroup
\end{table}

\noindent\textbf{Data scaling.}
We align the number of training epochs across all data configurations and
evaluate their performance. 

\begin{wrapfigure}[10]{r}{0.44\columnwidth}
    \vspace{-10pt}
    \centering
    \includegraphics[width=\linewidth]{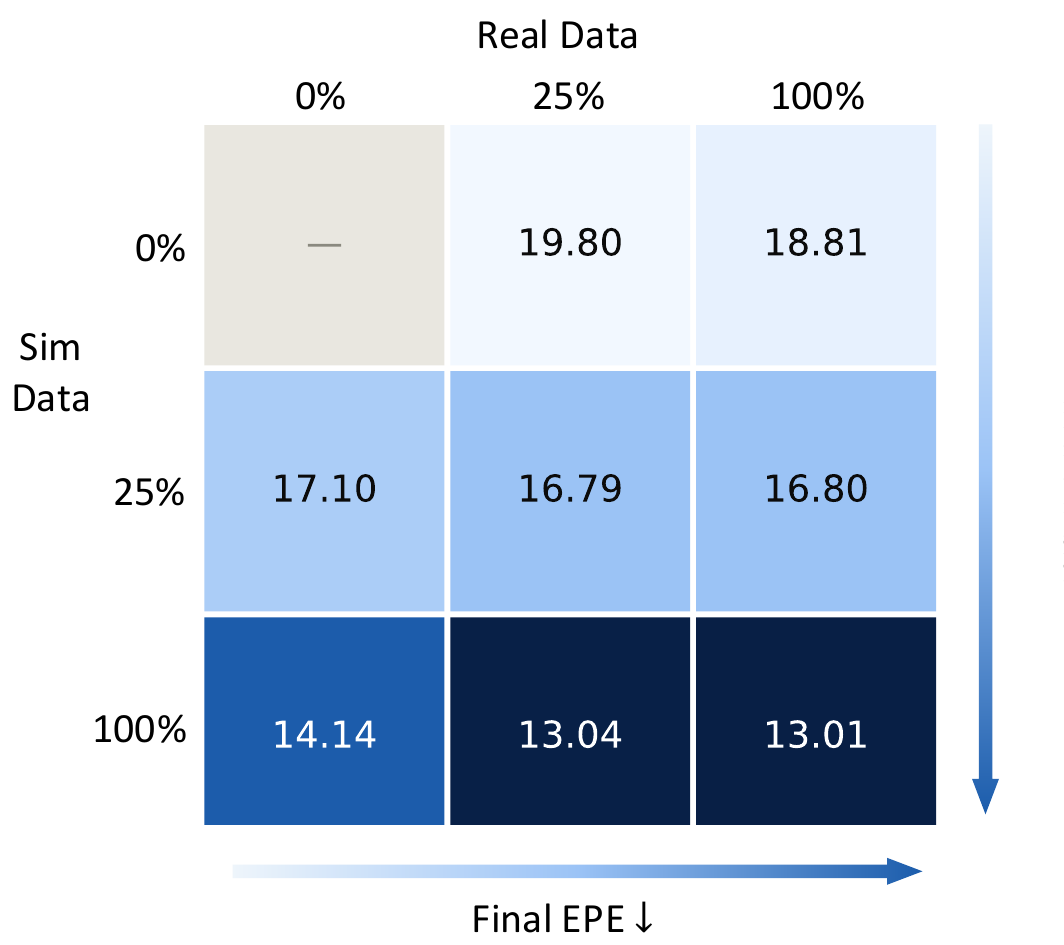}
    \vspace{-15pt}
    \caption{\textbf{B3. Data scaling.}}
    \label{fig:data_scaling}
    \vspace{-4pt}
\end{wrapfigure}

As shown in Fig.~\ref{fig:data_scaling}, prediction error generally decreases
as either simulated or real-world training data increases. Simulated data
provides the larger overall gain, while real-world data offers complementary
supervision for motions and observations encountered in real environments.
Overall, scaling either data source improves prediction accuracy, and combining the two yields the strongest performance.

\noindent\textbf{Effect of task-language conditioning.} Table~\ref{tab:language_ablation} examines whether task-specific language provides useful motion cues. Replacing placeholder captions with task descriptions reduces overall final-frame EPE from 13.76 to 13.01\,cm. The largest gains occur on AgiBot and SthV2, where EPE decreases by 1.53 and 2.21\,cm, respectively. These results confirm that task language provides meaningful conditioning for full-object motion prediction, particularly on heterogeneous real-world data.

\begin{table}[t]
\centering
\caption{\textbf{C1. Simulation execution results.}}
\vspace{-4pt}
\label{tab:sim_execution}

\begingroup
\small
\setlength{\tabcolsep}{2pt}
\renewcommand{\arraystretch}{1.0}

\begin{tabular*}{\linewidth}{
    @{\extracolsep{\fill}}lcccc@{}
}
\toprule
\multirow{2}{*}{Method}
& \multicolumn{2}{c}{\textbf{ManiSkill}}
& \multicolumn{2}{c}{\textbf{LIBERO}} \\[-2pt]
\cmidrule(lr){2-3}
\cmidrule(lr){4-5}
& PushCube
& PlaceSphere
& Insert
& Stack-Arrange \\[-2pt]
\midrule
TraceGen
& 5/10
& 4/10
& 7/10
& 7/10 \\
ObjectForesight
& 6/10
& 7/10
& 7/10
& 6/10 \\
\textbf{Ours}
& \textbf{10/10}
& \textbf{10/10}
& \textbf{9/10}
& \textbf{10/10} \\
\bottomrule
\end{tabular*}

\endgroup
\vspace{-15pt}
\end{table}
\begin{table}[t]
\centering
\caption{\textbf{C2. Real-world execution results.}}
\vspace{-5pt}
\label{tab:real_execution}

\begingroup
\small
\renewcommand{\arraystretch}{1.0}

\begin{tabular*}{\linewidth}{
    @{\extracolsep{\fill}}lcccc@{}
}
\toprule
\textbf{Method}
& \textbf{Banana}
& \textbf{Water}
& \textbf{Flashlight}
& \textbf{Cabbage} \\
\midrule
Dream2Flow
& 2/5 & 3/5 & \textbf{5/5} & 4/5 \\
Ours
& \textbf{4/5} & \textbf{4/5} & \textbf{5/5} & \textbf{5/5} \\
\bottomrule
\end{tabular*}

\endgroup
\vspace{-15pt}
\end{table}

\subsection{Full-Object Flow as an Execution Interface (Q4)}
\label{sec:exp_execution}

Tables~\ref{tab:sim_execution} and~\ref{tab:real_execution} report the
simulation and real-world execution results, respectively.
All methods share the same grasping, flow-to-action optimization, and
low-level execution pipeline, with the pre-grasp phase standardized. In simulation, we assess post-grasp execution through spatial trajectory
agreement with the ground truth, considering runs successful when they
meet a common coverage criterion within a fixed spatial tolerance.
Under this shared pipeline, our method generally achieves closer
agreement with the reference motion than the baselines.
These results support the viability of predicted full-object flow
as an execution interface between visual prediction and robot control.

\subsection{Environment and Embodiment Generalization (Q5)}
\label{sec:exp_generalization}

We evaluate transfer across environments and robot embodiments. Our method and the baselines are evaluated in ManiSkill and LIBERO
across different scenes and tasks, with results reported in
Table~\ref{tab:sim_execution}.
In the real world, we quantitatively evaluate a parallel gripper on
banana and flashlight pick-and-place tasks and water pouring, and a
dexterous hand on placing a cabbage onto a plate
(Table~\ref{tab:real_execution}). Example predictions are shown in
Fig.~\ref{fig:sim_and_real}. Together,
these results show that the object-flow interface can accommodate changes in
both environment and embodiment.
\section{Conclusion}

We introduced 3D full-object flow as an object-centric interface between predictive world models and downstream robot execution. By assigning dense trajectories to the complete target-object surface, this representation captures both visible and occluded object motion while remaining independent of a specific robot embodiment. We further built a large-scale full-object flow dataset from real-world videos and simulation, and trained a world model that predicts complete object motion from image-and-language inputs. Experiments demonstrate reliable annotation quality, improved flow prediction with scalable data, and effective downstream manipulation across simulation and real-world settings. These results suggest that full-object 3D motion prediction is a promising foundation for scalable and transferable robotic world models.
\section*{Acknowledgment}

This work was supported by the NUS Guangzhou Research Translation and Innovation Institute (GRTII)-Affiliated PhD Scholarship Program. We thank Zhehao Zhou, Jiongxu Chen, Zhuoman Liu, Yiyao Liu, Xinrui Qi, Minwen Liao, Yichen Guo, and the RoboScience data and simulation team for their valuable support and assistance.









\bibliographystyle{IEEEtran}
\bibliography{reference}

@inproceedings{finn2017deep,
  title={Deep visual foresight for planning robot motion},
  author={Finn, Chelsea and Levine, Sergey},
  booktitle={ICRA},
  year={2017},
}

@article{goalvla,
  title={Goal-VLA: Image-Generative VLMs as Object-Centric World Models Empowering Zero-shot Robot Manipulation},
  author={Chen, Haonan and Guo, Jingxiang and Wang, Bangjun and others},
  journal={arXiv preprint arXiv:2506.23919},
  year={2025}
}

@inproceedings{susie,
  title={Zero-shot robotic manipulation with pre-trained image-editing diffusion models},
  author={Black, Kevin and Nakamoto, Mitsuhiko and others},
  booktitle={ICLR},
  year={2024}
}

@article{wu2024ivideogpt,
  title={ivideogpt: Interactive videogpts are scalable world models},
  author={Wu, Jialong and Yin, Shaofeng and Feng, Ningya and He, Xu and Li, Dong and Hao, Jianye and Long, Mingsheng},
  journal={NeurIPS},
  year={2024}
}

@article{huang2024rekep,
  title={Rekep: Spatio-temporal reasoning of relational keypoint constraints for robotic manipulation},
  author={Huang, Wenlong and Wang, Chen and Li, Yunzhu and Zhang, Ruohan and Fei-Fei, Li},
  journal={arXiv preprint arXiv:2409.01652},
  year={2024}
}

@inproceedings{moka,
  title={Moka: Open-vocabulary robotic manipulation through mark-based visual prompting},
  author={Liu, Fangchen and Fang, Kuan and Abbeel, Pieter and others},
  booktitle={RSS},
  year={2024}
}

@article{objectforesight,
  title={ObjectForesight: Predicting Future 3D Object Trajectories from Human Videos},
  author={Soraki, Rustin and Bharadhwaj, Homanga and Farhadi, Ali and Mottaghi, Roozbeh},
  journal={arXiv preprint arXiv:2601.05237},
  year={2026}
}

@inproceedings{Im2flow2act,
  title={Flow as the cross-domain manipulation interface},
  author={Xu, Mengda and Xu, Zhenjia and Xu, Yinghao and Chi, Cheng and Wetzstein, Gordon and Veloso, Manuela and Song, Shuran},
  booktitle={CoRL},
  year={2024}
}

@article{dream2flow,
  title={Dream2Flow: Bridging Video Generation and Open-World Manipulation with 3D Object Flow},
  author={Dharmarajan, Karthik and others},
  journal={ICRA},
  year={2026}
}

@article{pointworld,
  title={PointWorld: Scaling 3D World Models for In-The-Wild Robotic Manipulation},
  author={Huang, Wenlong and Chao, Yu-Wei and Mousavian, Arsalan and others},
  journal={CVPR},
  year={2026}
}

@article{li2025novaflow,
  title={Novaflow: Zero-shot manipulation via actionable flow from generated videos},
  author={Li, Hongyu and Sun, Lingfeng and Hu, Yafei and Ta, Duy and Barry, Jennifer and Konidaris, George and Fu, Jiahui},
  journal={arXiv preprint arXiv:2510.08568},
  year={2025}
}

@article{damen2020epic,
  title={The epic-kitchens dataset: Collection, challenges and baselines},
  author={Damen, Dima and Doughty, Hazel and Farinella, Giovanni Maria and others},
  journal={TPAMI},
  year={2020},
  publisher={IEEE}
}

@inproceedings{goyal2017something,
  title={The" something something" video database for learning and evaluating visual common sense},
  author={Goyal, Raghav and Ebrahimi Kahou, Samira and Michalski, Vincent and others},
  booktitle={ICCV},
  year={2017}
}

@article{mahdisoltani2018effectiveness,
  title={On the effectiveness of task granularity for transfer learning},
  author={Mahdisoltani, Farzaneh and Berger, Guillaume and Gharbieh, Waseem and Fleet, David and Memisevic, Roland},
  journal={arXiv preprint arXiv:1804.09235},
  year={2018}
}

@inproceedings{grauman2022ego4d,
  title={Ego4d: Around the world in 3,000 hours of egocentric video},
  author={Grauman, Kristen and Westbury, Andrew and Byrne, Eugene and others},
  booktitle={CVPR},
  year={2022}
}

@inproceedings{liu2022hoi4d,
  title={Hoi4d: A 4d egocentric dataset for category-level human-object interaction},
  author={Liu, Yunze and Liu, Yun and Jiang, Che and Lyu, Kangbo and Wan, Weikang and Shen, Hao and Liang, Boqiang and Fu, Zhoujie and Wang, He and Yi, Li},
  booktitle={CVPR},
  year={2022}
}

@inproceedings{chao2021dexycb,
  title={Dexycb: A benchmark for capturing hand grasping of objects},
  author={Chao, Yu-Wei and Yang, Wei and Xiang, Yu and others},
  booktitle={CVPR},
  year={2021}
}

@inproceedings{fan2023arctic,
  title={ARCTIC: A dataset for dexterous bimanual hand-object manipulation},
  author={Fan, Zicong and Taheri, Omid and Tzionas, Dimitrios and others},
  booktitle={CVPR},
  year={2023}
}

@article{hot3d,
  title={{HOT3D}: Hand and Object Tracking in {3D} from Egocentric Multi-View Videos},
  author={Banerjee, Prithviraj and Shkodrani, Sindi and Moulon, Pierre and others},
  journal={CVPR},
  year={2025}
}

@inproceedings{o2024open,
  title={Open x-embodiment: Robotic learning datasets and rt-x models: Open x-embodiment collaboration 0},
  author={O’Neill, Abby and Rehman, Abdul and Maddukuri, Abhiram and others},
  booktitle={ICRA},
  year={2024},
}

@article{khazatsky2024droid,
  title={Droid: A large-scale in-the-wild robot manipulation dataset},
  author={Khazatsky, Alexander and Pertsch, Karl and Nair, Suraj and Balakrishna, Ashwin and others},
  journal={RSS},
  year={2024}
}

@article{bu2025agibot,
  title={Agibot world colosseo: A large-scale manipulation platform for scalable and intelligent embodied systems},
  author={Bu, Qingwen and Cai, Jisong and Chen, Li and others},
  journal={arXiv preprint arXiv:2503.06669},
  year={2025}
}

@inproceedings{sam2,
  title={Sam 2: Segment anything in images and videos},
  author={Ravi, Nikhila and Gabeur, Valentin and Hu, Yuan-Ting and others},
  booktitle={ICLR},
  year={2025}
}

@article{gemini2.5,
  title={Gemini 2.5: Pushing the frontier with advanced reasoning, multimodality, long context, and next generation agentic capabilities},
  author={Comanici, Gheorghe and Bieber, Eric and others},
  journal={arXiv preprint arXiv:2507.06261},
  year={2025}
}

@article{sam3d,
  title={Sam 3d: 3dfy anything in images},
  author={Chen, Xingyu and Chu, Fu-Jen and Gleize, Pierre and Liang, Kevin J and others},
  journal={arXiv preprint arXiv:2511.16624},
  year={2025}
}

@inproceedings{any6d,
  title={Any6D: Model-free 6D pose estimation of novel objects},
  author={Lee, Taeyeop and Wen, Bowen and Kang, Minjun and others},
  booktitle={CVPR},
  year={2025}
}

@inproceedings{wen2024foundationpose,
  title={Foundationpose: Unified 6d pose estimation and tracking of novel objects},
  author={Wen, Bowen and Yang, Wei and Kautz, Jan and others},
  booktitle={CVPR},
  year={2024}
}

@inproceedings{ptv3,
  title={Point transformer v3: Simpler faster stronger},
  author={Wu, Xiaoyang and Jiang, Li and Wang, Peng-Shuai and others},
  booktitle={CVPR},
  year={2024}
}

@inproceedings{gao2025flip,
  title={Flip: Flow-centric generative planning as general-purpose manipulation world model},
  author={Gao, Chongkai and Zhang, Haozhuo and others},
  booktitle={ICLR},
  year={2025}
}

@article{tao2024maniskill3,
  title={Maniskill3: Gpu parallelized robotics simulation and rendering for generalizable embodied ai},
  author={Tao, Stone and Xiang, Fanbo and Shukla, Arth and Qin, Yuzhe and Hinrichsen, Xander and others},
  journal={arXiv preprint arXiv:2410.00425},
  year={2024}
}

@inproceedings{chen2025metafold,
  title={Metafold: Language-guided multi-category garment folding framework via trajectory generation and foundation model},
  author={Chen, Haonan and Li, Junxiao and Wu, Ruihai and others},
  booktitle={IROS},
  year={2025},
}

@inproceedings{yu2020metaworld,
  title={Meta-world: A benchmark and evaluation for multi-task and meta reinforcement learning},
  author={Yu, Tianhe and Quillen, Deirdre and He, Zhanpeng and Julian, Ryan and Hausman, Karol and Finn, Chelsea and Levine, Sergey},
  booktitle={CoRL},
  year={2020},
  organization={PMLR}
}

@article{james2020rlbench,
  title={Rlbench: The robot learning benchmark \& learning environment},
  author={James, Stephen and Ma, Zicong and Arrojo, David Rovick and Davison, Andrew J},
  journal={IEEE Robotics and Automation Letters},
  volume={5},
  number={2},
  pages={3019--3026},
  year={2020},
  publisher={IEEE}
}

@article{nasiriany2024robocasa,
  title={Robocasa: Large-scale simulation of everyday tasks for generalist robots},
  author={Nasiriany, Soroush and Maddukuri, Abhiram and Zhang, Lance and Parikh, Adeet and Lo, Aaron and Joshi, Abhishek and others},
  journal={arXiv preprint arXiv:2406.02523},
  year={2024}
}

@article{chen2025robotwin2,
  title={Robotwin 2.0: A scalable data generator and benchmark with strong domain randomization for robust bimanual robotic manipulation},
  author={Chen, Tianxing and Chen, Zanxin and others},
  journal={arXiv preprint arXiv:2506.18088},
  year={2025}
}

@article{mikasa,
  title={Memory, benchmark \& robots: A benchmark for solving complex tasks with reinforcement learning},
  author={Cherepanov, Egor and Kachaev, Nikita and Kovalev, Alexey K and Panov, Aleksandr I},
  journal={arXiv preprint arXiv:2502.10550},
  year={2025}
}

@inproceedings{veicht2024geocalib,
  title={Geocalib: Learning single-image calibration with geometric optimization},
  author={Veicht, Alexander and Sarlin, Paul-Edouard and Lindenberger, Philipp and others},
  booktitle={ECCV},
  year={2024},
}

@article{tro,
  title={T (R, O) Grasp: Efficient Graph Diffusion of Robot-Object Spatial Transformation for Cross-Embodiment Dexterous Grasping},
  author={Fei, Xin and Xu, Zhixuan and Fang, Huaicong and others},
  journal={arXiv preprint arXiv:2510.12724},
  year={2025}
}

@article{liu2023libero,
  title={Libero: Benchmarking knowledge transfer for lifelong robot learning},
  author={Liu, Bo and Zhu, Yifeng and Gao, Chongkai and others},
  journal={NeurIPS},
  year={2023}
}

@article{gao2026vla,
  title={Vla-os: Structuring and dissecting planning representations and paradigms in vision-language-action models},
  author={Gao, Chongkai and Liu, Zixuan and others},
  journal={NeurIPS},
  year={2025}
}

@article{openvla,
  title={Openvla: An open-source vision-language-action model},
  author={Kim, Moo Jin and Pertsch, Karl and others},
  journal={arXiv preprint arXiv:2406.09246},
  year={2024}
}

@inproceedings{rt2,
  title={Rt-2: Vision-language-action models transfer web knowledge to robotic control},
  author={Zitkovich, Brianna and Yu, Tianhe and Xu, Sichun and others},
  booktitle={Conference on Robot Learning},
  pages={2165--2183},
  year={2023},
  organization={PMLR}
}

@article{hou2026world,
  title={World Model for Robot Learning: A Comprehensive Survey},
  author={Hou, Bohan and Li, Gen and Jia, Jindou and others},
  journal={arXiv preprint arXiv:2605.00080},
  year={2026}
}

@article{firoozi2025foundation,
  title={Foundation models in robotics: Applications, challenges, and the future},
  author={Firoozi, Roya and Tucker, Johnathan and Tian, Stephen and others},
  journal={IJRR},
  year={2025},
  publisher={Sage Publications Sage UK: London, England}
}

@article{wang2026vision,
  title={Vision-Language-Action in Robotics: A Survey of Datasets, Benchmarks, and Data Engines},
  author={Wang, Ziyao and Wang, Bingying and Zhang, Hanrong and others},
  journal={arXiv preprint arXiv:2604.23001},
  year={2026}
}

@article{ebert2018visual,
  title={Visual foresight: Model-based deep reinforcement learning for vision-based robotic control},
  author={Ebert, Frederik and Finn, Chelsea and Dasari, Sudeep and Xie, Annie and Lee, Alex and Levine, Sergey},
  journal={arXiv preprint arXiv:1812.00568},
  year={2018}
}

@article{videomanip,
  title={Dexterous manipulation policies from rgb human videos via 3d hand-object trajectory reconstruction},
  author={Chen, Hongyi and Dong, Tony and Wu, Tiancheng and others},
  journal={arXiv preprint arXiv:2602.09013},
  year={2026}
}

@article{sec,
  title={Advancing Complex Video Object Segmentation via Progressive Concept Construction},
  author={Zhang, Zhixiong and Ding, Shuangrui and Dong, Xiaoyi and others},
  journal={arXiv preprint arXiv:2507.15852},
  year={2025}
}

@article{track4world,
  title={Track4world: Feedforward world-centric dense 3d tracking of all pixels},
  author={Lu, Jiahao and Xu, Jiayi and others},
  journal={arXiv preprint arXiv:2603.02573},
  year={2026}
}

@article{depthanything3,
  title={Depth anything 3: Recovering the visual space from any views},
  author={Lin, Haotong and Chen, Sili and Liew, Junhao and Chen, Donny Y and Li, Zhenyu and Shi, Guang and Feng, Jiashi and Kang, Bingyi},
  journal={arXiv preprint arXiv:2511.10647},
  year={2025}
}

@inproceedings{lee2026tracegen,
  title={Tracegen: World modeling in 3d trace space enables learning from cross-embodiment videos},
  author={Lee, Seungjae and Jung, Yoonkyo and Chun, Inkook and others},
  booktitle={CVPR},
  year={2026}
}

@misc{bai2025qwen25vltechnicalreport,
      title={Qwen2.5-VL Technical Report}, 
      author={Shuai Bai and Keqin Chen and Xuejing Liu and others},
      year={2025},
      eprint={2502.13923},
      archivePrefix={arXiv},
      primaryClass={cs.CV},
      url={https://arxiv.org/abs/2502.13923}, 
}

\useRomanappendicesfalse
\appendices

\label{sec:appendix}

\section{Annotation of Real-World Videos}
\label{app:dataset_pipeline}

Fig.~\ref{fig:data_pipeline} illustrates our automatic annotation pipeline, which converts uncalibrated RGB videos into time-indexed 3D full-object flow supervision. For each action clip, the pipeline produces a task description $\mathcal{L}$ and object trajectories $\mathcal{P}^{\mathrm{obj}}_{1:T}$. The pipeline requires only RGB videos, without robot action annotations, depth measurements, camera calibration, or pre-scanned object meshes, enabling annotation of heterogeneous robotic and human manipulation videos captured from different viewpoints.

\subsection{VLM-based Annotation and Action Segmentation}
Given a raw manipulation video, we use Gemini 2.5 Pro~\cite{gemini2.5} to identify manipulation events and divide the video into atomic action clips. Each clip represents a single contact-and-separation interaction between the actor and the manipulated object. For each clip, the VLM outputs the target object category, a natural language task description $\mathcal{L}$, and the temporal boundaries of the action. These annotations guide subsequent object-centric processing.

\subsection{4D Scene Reconstruction and Point Tracking}
Given the RGB frames of an action clip, we apply Track4World~\cite{track4world}, with Depth Anything 3 (DA3)~\cite{depthanything3} as its backbone, to estimate scene geometry and camera motion. The module outputs per-frame depth estimates, camera intrinsics, camera poses, and temporally tracked scene points.

We denote the camera coordinate system at time $t$ by $C_t$ and use the first-frame camera $C_1$ as the reference coordinate system. In particular, Track4World provides the camera transformation $\mathbf{T}^{C_1}_{C_t}$, allowing reconstructed points from different frames to be expressed in a common coordinate system. The recovered geometry and camera information provide the spatial reference for subsequent object pose estimation and full-object flow extraction.

\subsection{Target Object Segmentation}
Given the target object category from the VLM annotations and the RGB action clip, we localize the actively manipulated object using a two-stage strategy. First, the VLM proposes and scores target-object bounding boxes across candidate frames. The highest-scoring proposal identifies the target instance.

We then use SeC~\cite{sec} to segment the localized object and propagate its mask throughout the clip, producing a sequence of binary object masks $\{M_t\}_{t=1}^{T}$. These masks isolate the manipulated object from the surrounding scene and provide object-specific visual evidence for reconstruction and pose tracking. The explicit localization stage helps disambiguate the manipulated instance when multiple objects share the same category.

\subsection{Object Shape Reconstruction}
Given the RGB observations and target masks, SAM 3D~\cite{sam3d} reconstructs a complete object mesh $\mathcal{M}^{\mathrm{obj}}$, including inferred geometry for occluded or unobserved regions.

We denote the canonical coordinate frame of the reconstructed object by $O$. The mesh provides a persistent object-centric geometric representation from which surface points can be sampled and tracked over time.

\subsection{Object Pose Estimation and Tracking}
Given the reconstructed mesh, target-object observations, and estimated scene geometry, we use Any6D~\cite{any6d} to estimate the object's initial 6-DoF pose and refine its geometric scale. This produces an initialized object mesh and its pose $\mathbf{T}^{C_1}_{O,1}$ relative to the first-frame camera.

We then initialize FoundationPose~\cite{wen2024foundationpose} with the reconstructed object geometry and initial pose. Using the RGB observations, estimated depth, camera intrinsics, and object masks, FoundationPose tracks the object throughout the clip, producing the time-indexed pose sequence
$\{\mathbf{T}^{C_t}_{O,t}\}_{t=1}^{T}$.
Each transformation maps points from the canonical object frame $O$ to the corresponding camera frame $C_t$.

\subsection{Full-Object Flow Extraction}
Given the reconstructed object mesh and its estimated pose sequence, we uniformly sample $N^{\mathrm{obj}}=512$ surface points
$\{\bar{\mathbf{x}}_i\}_{i=1}^{N^{\mathrm{obj}}}$
in the canonical object frame $O$. These points are propagated through the estimated pose sequence, preserving their correspondences across time.

Let $\mathbf{T}^{A}_{B}\in SE(3)$ denote a transformation mapping coordinates from frame $B$ to frame $A$. For each object point, we compose the tracked object pose with the camera transformation estimated by Track4World:
\begin{equation}
    \mathbf{p}_{i,t}
=
\mathbf{T}^{C_1}_{C_t}
\mathbf{T}^{C_t}_{O,t}
\bar{\mathbf{x}}_i,
\quad
i=1,\ldots,N^{\mathrm{obj}},\quad
t=1,\ldots,T,
\end{equation}

where homogeneous coordinates are omitted for clarity.

This maps every sampled object point into the common first-frame camera coordinate system, yielding
\begin{equation}
    \mathcal{P}^{\mathrm{obj}}_{1:T}
=
\{\mathbf{p}_{i,t}\in\mathbb{R}^{3}\}
_{i=1,t=1}^{N^{\mathrm{obj}},T}.
\end{equation}

Because the sampled points cover the reconstructed complete object surface, their trajectories describe both visible and initially occluded regions, rather than only the points observed in the input frames.


\section{Dataset Details}
\label{app:dataset_details}

\subsection{Sources and Composition}

Our dataset comprises 1,210.1 hours of manipulation data measured at a
standardized 16 FPS: 454.5 hours of simulated trajectories and 755.6 hours
of automatically annotated real-world videos.
Table~\ref{tab:dataset_composition} provides the source-level breakdown,
including sample counts and average frame counts before fixed-length
sampling.

For the simulation subset, we aggregate 198,947 trajectories spanning
251 tasks from seven datasets: ManiSkill 3~\cite{tao2024maniskill3},
MetaFold~\cite{chen2025metafold}, Meta-World~\cite{yu2020metaworld},
MIKASA~\cite{mikasa}, RLBench~\cite{james2020rlbench},
RoboCasa~\cite{nasiriany2024robocasa}, and
RoboTwin 2.0~\cite{chen2025robotwin2}.
MetaFold contributes deformable-object manipulation data.

For real-world data, we collect videos with diverse viewpoints and
embodiments from three sources: robot demonstrations from
AgiBot World~\cite{bu2025agibot}, third-person human demonstrations from
Something-Something V2~\cite{goyal2017something}, and egocentric human
demonstrations from EPIC-KITCHENS~\cite{damen2020epic}.
After automatic annotation and quality filtering, we retain 289,976 clips
totaling 755.6 hours.

\begin{table}[t]
    \centering
    \caption{\textbf{Dataset composition.}
    Average frame counts are measured before fixed-length sampling.
    Sth-V2 denotes Something-Something V2.}
    \label{tab:dataset_composition}
    \small
    \setlength{\tabcolsep}{3pt}
    \begin{tabular*}{\columnwidth}
        {@{\extracolsep{\fill}}lrrr@{}}
        \toprule
        Source & Samples & \shortstack{Avg.\\frames} & Hours \\
        \midrule
        \multicolumn{4}{@{}l}{\textit{Simulation}} \\
        ManiSkill 3  & 8,421   & 111.9 & 16.4 \\
        MIKASA       & 4,599   & 91.0  & 7.3 \\
        Meta-World   & 52,500  & 76.5  & 69.7 \\
        MetaFold     & 2,014   & 101.0 & 3.5 \\
        RLBench      & 3,386   & 156.2 & 9.2 \\
        RoboCasa     & 12,230  & 234.5 & 49.8 \\
        RoboTwin 2.0 & 115,797 & 148.5 & 298.6 \\
        Subtotal     & 198,947 & 131.6 & 454.5 \\
        \midrule
        \multicolumn{4}{@{}l}{\textit{Real world}} \\
        AgiBot World  & 244,191 & 169.5 & 718.6 \\
        Sth-V2          & 44,259  & 44.4  & 34.1 \\
        EPIC-KITCHENS & 1,526   & 108.1 & 2.9 \\
        Subtotal      & 289,976 & 150.1 & 755.6 \\
        \midrule
        \textbf{Total}
        & \textbf{488,923}
        & \textbf{142.5}
        & \textbf{1,210.1} \\
        \bottomrule
    \end{tabular*}
\end{table}

\subsection{Real-World Quality Subsets}

We organize real-world annotations into two nested quality subsets
using the reprojection-IoU score. The Silver set contains all retained
clips with scores of at least 0.50. The Golden set is its higher-quality
subset, containing clips with scores of at least 0.70.

Table~\ref{tab:real_quality_subsets} reports their source-level
statistics. The Silver set comprises the full 755.6-hour real-world
collection, while Golden contains 120,269 clips totaling 288.7 hours.
The main model is trained on the simulation data together with the
Golden real-world subset.

\begin{table}[t]
    \centering
    \caption{\textbf{Real-world quality subsets.}
    Golden is a subset of Silver.
    Average frame counts are measured before fixed-length sampling.}
    \label{tab:real_quality_subsets}
    \small
    \setlength{\tabcolsep}{3pt}
    \begin{tabular*}{\columnwidth}
        {@{\extracolsep{\fill}}lrrr@{}}
        \toprule
        Source & Samples & \shortstack{Avg.\\frames} & Hours \\
        \midrule
        \multicolumn{4}{@{}l}{\textit{Golden}} \\
        AgiBot World  & 99,360  & 158.0 & 272.5 \\
        Sth-V2          & 20,573  & 43.7  & 15.6 \\
        EPIC-KITCHENS & 336     & 101.2 & 0.6 \\
        Subtotal      & 120,269 & 138.3 & 288.7 \\
        \midrule
        \multicolumn{4}{@{}l}{\textit{Silver (including Golden)}} \\
        AgiBot World  & 244,191 & 169.5 & 718.6 \\
        Sth-V2          & 44,259  & 44.4  & 34.1 \\
        EPIC-KITCHENS & 1,526   & 108.1 & 2.9 \\
        Subtotal      & 289,976 & 150.1 & 755.6 \\
        \bottomrule
    \end{tabular*}
\end{table}

\subsection{Motion Annotations and Sequence Organization}

For rigid simulated objects, we construct full-object trajectories
by sampling object surfaces and transforming the points using
simulator-provided pose sequences. For deformable objects,
trajectories follow corresponding mesh vertices across frames.
Real-world trajectories are recovered through the reconstruction
and pose-tracking pipeline described in Sec.~\ref{sec:data_pipeline}.
Both pathways produce dense 3D trajectories over the complete
target-object surface.

We organize each sequence into 49 sampled frames by subsampling
longer sequences and repeating the final frame of shorter ones.
This provides an initial observation and 48 subsequent prediction
steps. The average frame counts in
Tables~\ref{tab:dataset_composition} and~\ref{tab:real_quality_subsets}
describe sequences before this fixed-length sampling.

\begin{figure*}[t]
    \centering
    \includegraphics[width=\textwidth]{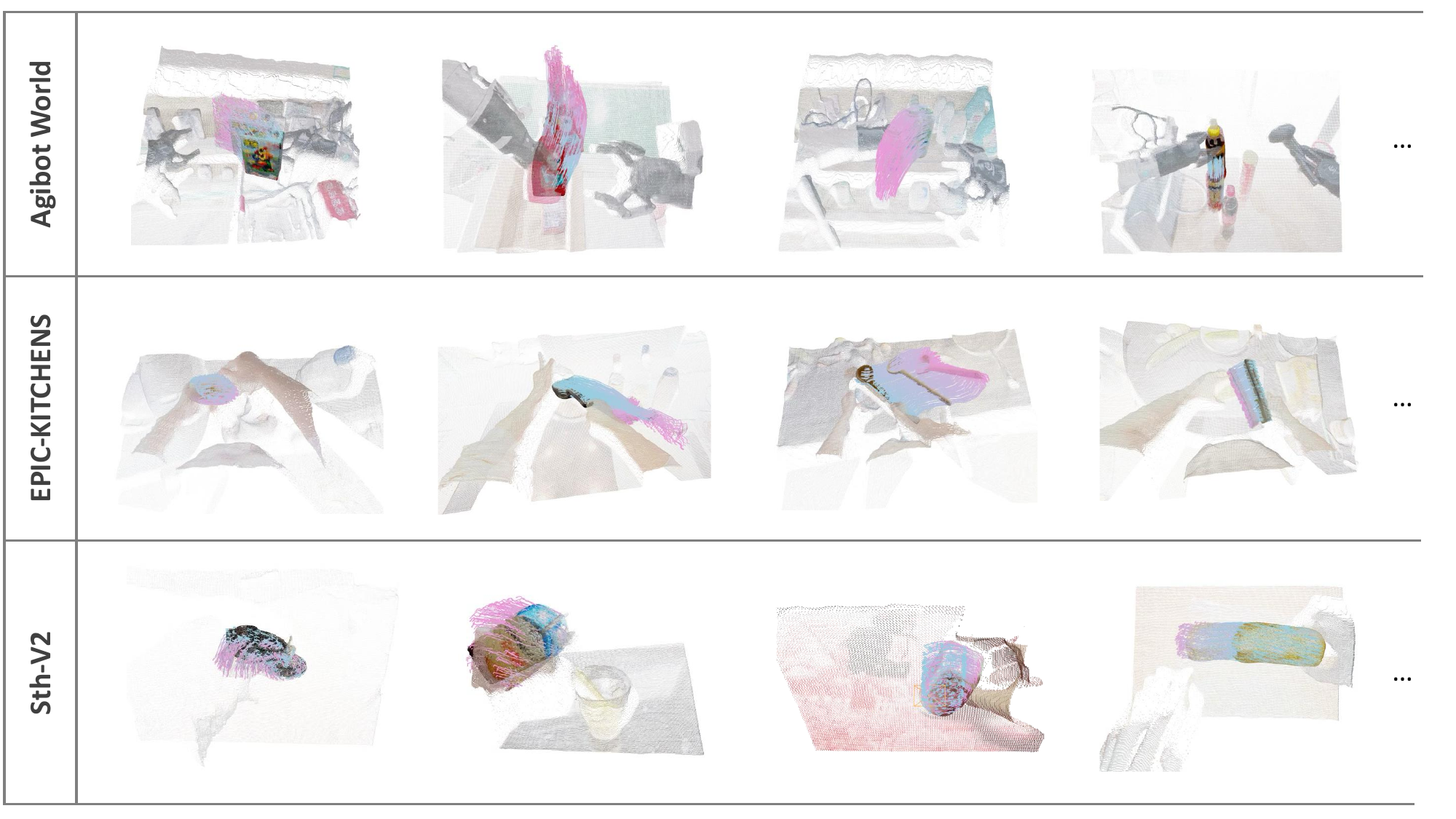}
    \caption{\textbf{Visualization of real-world full-object flow annotations.}
    Examples from AgiBot World~\cite{bu2025agibot}, EPIC-KITCHENS~\cite{damen2020epic}, and Something-Something V2~\cite{goyal2017something}.
    Colored trajectories represent the estimated motion of points on the reconstructed complete object surface, with colors progressing from red at earlier timesteps to blue at later timesteps.
    Each example corresponds to a different manipulation clip processed by our automatic annotation pipeline.}
    \label{fig:real_anno}
\end{figure*}

\subsection{Visualization of Real-World Annotations}

Fig.~\ref{fig:real_anno} presents qualitative examples of automatically annotated full-object flow from three real-world video sources: AgiBot World~\cite{bu2025agibot}, EPIC-KITCHENS~\cite{damen2020epic}, and Something-Something V2~\cite{goyal2017something}. These sources provide broad coverage of object categories, manipulation tasks, and visual environments, encompassing both robotic and human demonstrations captured from different viewpoints. Our annotation pipeline converts these heterogeneous observations into a unified 3D representation, recovering trajectories over the reconstructed complete object surface in a common coordinate system. These examples illustrate the diversity of real-world manipulation data that can be incorporated into full-object flow supervision without requiring robot-specific action annotations or calibrated sensing setups.

\section{Model and Training Details}
\label{app:model_training}

We provide additional implementation details for the full-object flow
predictor introduced in Sec.~\ref{sec:model}.

\subsection{Multimodal Observation Encoding}

\noindent\textbf{Input representation.}
The initial object surface
$\mathbf{P}^{o}\in\mathbb{R}^{N_o\times 3}$ consists of mesh-sampled
points, while $\mathbf{P}^{s}\in\mathbb{R}^{N_s\times 3}$ represents
sampled scene geometry. We represent the target trajectories as
displacements $\mathbf{F}\in\mathbb{R}^{T\times N_o\times 3}$ from
the initial object surface, with $T=48$. The geometric inputs and
displacement targets are expressed in a common gravity-aligned
coordinate frame. Image-feature sampling uses the original
camera-frame coordinates and camera intrinsics.

\noindent\textbf{Visual-language pathway.}
We use a frozen Qwen2.5-VL-7B-Instruct encoder to extract task-conditioned
visual features from the initial RGB image and language instruction.
The instruction precedes the image in the input sequence, allowing
image tokens to incorporate task information through causal attention.
We reshape the image-token hidden states into a spatial feature grid
and obtain point-aligned features $\mathbf{f}^{\mathrm{vlm}}_i$
through image projection and bilinear sampling.
A separate text-only forward pass provides language tokens for
subsequent cross-attention. Both feature types are projected
into a shared 512-dimensional representation.

\noindent\textbf{Geometric pathway.}
We jointly encode the object and scene point clouds using
PTv3~\cite{ptv3}, with point coordinates as input features.
The resulting object and scene features are projected into the
shared feature space. As described in Sec.~\ref{sec:model},
the object tokens $\mathbf{H}^{o}$ combine coordinate embeddings,
point-aligned visual-language features, and geometric features.
The scene features $\mathbf{H}^{s}$ and language tokens
$\mathbf{H}^{l}$ provide additional context for trajectory prediction.

\subsection{Conditional Trajectory Generation}

We use the displacement trajectory $\mathbf{F}$ as the clean target
$\mathbf{x}_1$ in the linear noising formulation of
Sec.~\ref{sec:model}. For each object point, the noisy trajectory
$\mathbf{x}_t$ is flattened and projected into the object-token
feature space. The noise level $t$ is encoded using sinusoidal
embeddings and used to condition the network through adaptive
normalization.

The conditioned object tokens are processed by self-attention
and a Transformer decoder, where object tokens serve as queries
and scene and language tokens provide cross-attention context.
The decoder features are combined with the input through a skip
connection and passed to a flow prediction head, which also
incorporates the noisy trajectory. The network outputs
$\hat{\mathbf{x}}_1\in\mathbb{R}^{T\times N_o\times 3}$,
jointly predicting the complete displacement trajectory of
each object point.

We train the predictor using the mean-squared error between
$\hat{\mathbf{x}}_1$ and $\mathbf{x}_1$, averaged over points,
timesteps, and coordinate dimensions. The same objective is
applied to rigid and deformable trajectories.

\begin{table*}[t]
    \centering
    \caption{\textbf{Component ablations.}
    EPE is reported in centimeters, while Acc@5 and Acc@10 denote pooled per-point accuracies (\%). All configurations use the same evaluation set.}
    \label{tab:component_ablations}
    \small
    \setlength{\tabcolsep}{12pt}
    \renewcommand{\arraystretch}{1.1}
    \begin{tabular}{lcccccc}
        \toprule
        & \multicolumn{3}{c}{All Frames}
        & \multicolumn{3}{c}{Final Frame} \\
        \cmidrule(lr){2-4}
        \cmidrule(lr){5-7}
        Configuration
        & EPE $\downarrow$
        & Acc@5 $\uparrow$
        & Acc@10 $\uparrow$
        & EPE $\downarrow$
        & Acc@5 $\uparrow$
        & Acc@10 $\uparrow$ \\
        \midrule
        Full model
        & \textbf{6.84}
        & \textbf{57.5}
        & \textbf{76.5}
        & \textbf{12.69}
        & \textbf{19.6}
        & \textbf{52.0} \\
        3D only (w/o VLM)
        & 8.61 & 49.6 & 67.6
        & 16.90 & 9.2 & 31.5 \\
        Separate PTv3
        & 7.24 & 56.6 & 74.4
        & 13.46 & 17.0 & 45.9 \\
        \bottomrule
    \end{tabular}
\end{table*}

\subsection{Data Mixture and Optimization}

We train the model using all seven simulation sources and
the Golden subsets of AgiBot World, Something-Something V2,
and EPIC-KITCHENS. Each action segment is represented by 49 sampled frames, comprising
an initial observation and $T=48$ future frames over normalized
action progress.
The same gravity alignment is applied to the geometric
observation and trajectory target. We reserve 5\% of the
trajectories for validation using a deterministic sample-level
hash split.

The main model is trained for three epochs on eight NVIDIA A100
GPUs with a global batch size of 64. Data-scaling experiments
use one training epoch per configuration. The visual-language
encoder remains frozen, while the geometric encoder and
trajectory prediction modules are optimized jointly.

\subsection{Trajectory Sampling}

At inference time, we initialize the displacement trajectory
with Gaussian noise and use a uniformly spaced sampling
schedule from $t=1$ to $t=0$. At each step, the network
predicts the clean trajectory $\hat{\mathbf{x}}_1$.
For a subsequent noise level $s<t$, we recover the noise
estimate and update the trajectory as
\begin{align}
    \hat{\mathbf{x}}_0
    &=
    \frac{\mathbf{x}_t-(1-t)\hat{\mathbf{x}}_1}{t},\\
    \mathbf{x}_s
    &=
    (1-s)\hat{\mathbf{x}}_1+s\hat{\mathbf{x}}_0.
\end{align}
The final update at $s=0$ produces the predicted displacement
trajectory $\hat{\mathbf{F}}$. The corresponding full-object
flow is obtained by adding these displacements to the initial
object surface:
\begin{equation}
    \hat{\mathbf{P}}^{o}_t
    =
    \mathbf{P}^{o}+\hat{\mathbf{F}}_t.
\end{equation}

We use one sampling step for standard flow-prediction evaluation
and five steps for downstream robot execution. Five-step flow
generation takes approximately 1.08 seconds per trajectory
on a single A100 GPU.

\section{Component Ablations}
\label{sec:component_ablations}

We further evaluate two architectural choices in our full-object flow predictor: visual-language conditioning and joint object-scene geometric encoding. All configurations are evaluated on the same fixed supplementary evaluation set, comprising 2,048 samples. Table~\ref{tab:component_ablations} reports the mean endpoint error (EPE) and per-point accuracy at 5 cm and 10 cm thresholds.

\noindent\textbf{Visual-language conditioning.}
We examine the contribution of visual-language conditioning by removing both the visual features extracted by Qwen2.5-VL and the language-token context, leaving a predictor conditioned only on 3D object and scene geometry. As shown in Table~\ref{tab:component_ablations}, this increases final-frame EPE by 4.21 cm and reduces Acc@10 by 20.5 percentage points. While 3D geometry provides spatial information, visual-language conditioning supplies complementary semantic and task context for determining the intended object motion. The results highlight its importance for task-dependent full-object flow prediction.

\noindent\textbf{Joint object-scene geometric encoding.}
We further investigate whether object and scene geometry should be encoded jointly or independently. Unlike the full model, the separate PTv3 variant processes object and scene point clouds independently before integrating their features in the decoder. Joint encoding achieves better overall prediction accuracy, reducing final-frame EPE from 13.46 to 12.69 cm and improving Acc@10 from 45.9\% to 52.0\%. This suggests that incorporating surrounding scene geometry during initial object feature extraction enables more effective object-scene interactions and improves full-object flow prediction.

\end{document}